\documentclass[10pt,twocolumn,letterpaper]{article}

\usepackage{cvpr}      

\usepackage{placeins}
\usepackage{multirow}
\usepackage{array}

\definecolor{cvprblue}{rgb}{0.21,0.49,0.74}
\usepackage[pagebackref,breaklinks,colorlinks,allcolors=cvprblue]{hyperref}

\usepackage{multicol}
\usepackage{caption}
\usepackage{dsfont}
\usepackage{array}
\usepackage{cuted}

\def\confName{CVPR}
\def\confYear{2027}

\title{Beyond Head Importance: Collective Computation in Video Transformers}

\author{%
Yi Yang\textsuperscript{1} \quad
Siyuan Liu\textsuperscript{1} \quad
Bingbing Nie\textsuperscript{1,2}\thanks{%
Correspondence to: \texttt{nbb@tsinghua.edu.cn}}\\[4pt]
{\fontsize{10.5}{12.5}\selectfont
\textsuperscript{1}School of Vehicle and Mobility, Tsinghua University, China}\\
{\fontsize{10.5}{12.5}\selectfont
\textsuperscript{2}State Key Laboratory of Intelligent Green Vehicle and Mobility, Tsinghua University, China}
}

\begin{document}
\maketitle
\begin{strip}
    \centering
    \vspace*{-14mm}
    \includegraphics[width=\textwidth]{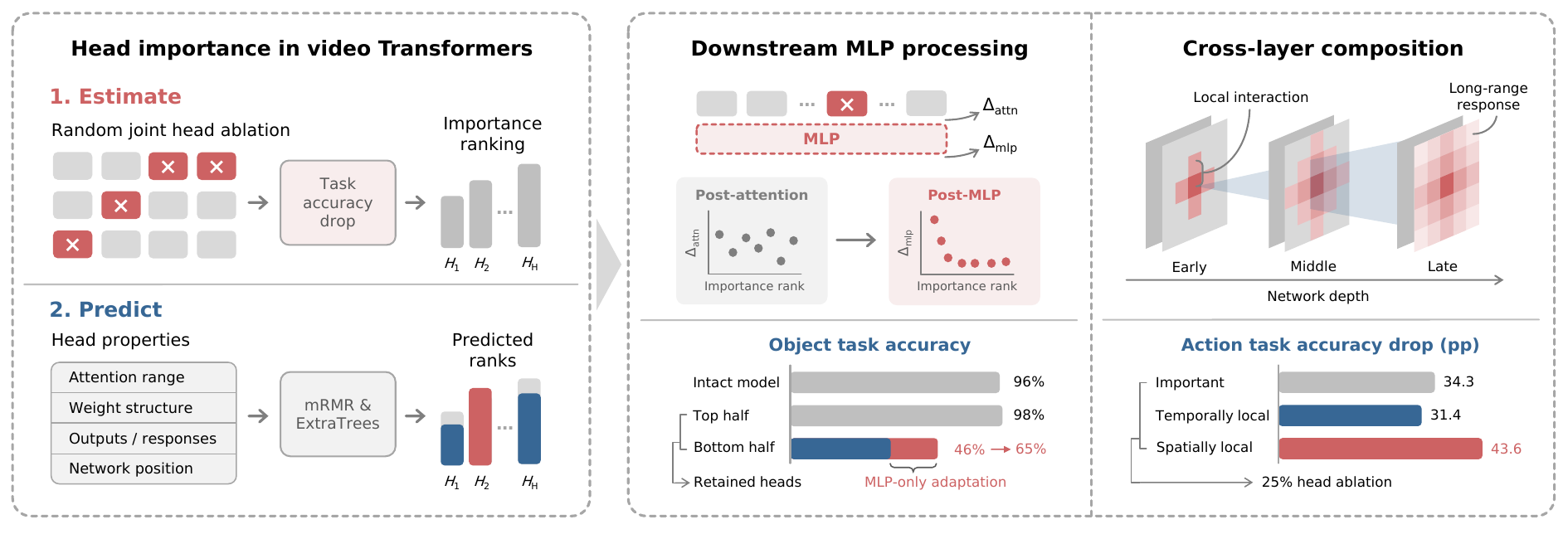}
\end{strip}

\begin{abstract}
Attention heads in video Transformers vary greatly in functional importance.
Understanding their contributions requires examining not only which heads are important, but also how they participate in collective computation within the network.
We study this collective computation from two perspectives: the downstream use of head outputs and the composition of interactions across layers.
Ablation analyses of four video Transformers show that high-importance heads identified on a single task also support other video understanding tasks and help preserve the representational structure that distinguishes different videos.
Representational responses to head ablation correlate weakly with head importance after the attention module, but more strongly after the MLP in the same block. Adapting this downstream processing through MLP-only fine-tuning further improves the use of lower-importance head outputs.
Meanwhile, through composition across layers, sets of heads with shorter spatiotemporal attention distances support stronger perturbation responses at distant tokens than sets with longer distances.
Their joint ablation causes even greater performance loss than ablating equally sized sets of high-importance heads, revealing collective contributions not fully captured by individual importance rankings.
These findings highlight collective computation as a key perspective for understanding attention head function: functional contributions depend on both how head outputs are used downstream and how local interactions compose across layers.
\end{abstract}

\section{Introduction}
\label{sec:intro}

Understanding how deep neural networks form and use information through internal computation is an important goal of model interpretability research~\cite{conmy2023towards,geiger2025causal,olah2020zoom}. Video understanding requires models to integrate information across space and time to recognize visual content and capture the dynamics of actions, motion, and events~\cite{zhou2018temporal,bertasius2021space,wu2022memvit}. Video Transformers use multi-head self-attention to enable spatiotemporal interactions and progressively build visual representations across layers~\cite{vaswani2017attention,arnab2021vivit,fan2021Multiscale,Zhang2021VidTr}. Representation probing and attention analysis can reveal the semantic and dynamic information encoded by these models, but understanding how representations are formed and maintained also requires examining their internal computations~\cite{kowal2024understanding,joseph2026interpreting,jun2026imamap,elazar2021amnesic}. In particular, how attention heads contribute to collective computation in video Transformers remains unclear.

The functional importance of attention heads in Transformers is highly uneven: ablating many low-importance heads has little effect on task performance, whereas ablating just a few high-importance heads can substantially degrade performance~\cite{michel2019sixteen,voita2019analyzing,kowal2024understanding,palenzuela2025importance}. Ablation analysis can identify the key heads on which a model relies, but offers limited insight into the computations underlying their functional contributions~\cite{li2024optimal,jacovi2020towards}. Individual head importance may also fail to fully capture the collective contribution of multiple heads working together~\cite{budhraja2020weak,nam2025causal}.

We therefore ask: \textbf{How do attention heads participate in collective computation in video Transformers?} We approach this question from two perspectives. First, head outputs are processed by MLPs and subsequent network layers, making their contributions dependent on how downstream computation uses these signals. Second, a head's attention range describes only a single interaction, whereas local interactions can combine across layers to produce broader spatiotemporal effects. We first use cross-task ablation analysis to identify heads that support multiple tasks, then examine how downstream processing and cross-layer composition shape their functional contributions.

We conduct attention head ablation analysis in four video Transformers: V-JEPA~\cite{bardes2024revisiting} and V-JEPA 2~\cite{assran2025vjepa2}, which are trained through feature prediction, and VideoMAE~\cite{tong2022videomae} and VideoMAE V2~\cite{wang2023videomaev2}, which are trained through masked reconstruction. We then focus on V-JEPA to investigate collective computation through downstream processing and cross-layer interactions. Our main contributions are as follows:
\begin{itemize}
    \item \textbf{We characterize the cross-task importance and representational roles of attention heads.} High-importance head sets identified on a single task also support other video understanding tasks and help preserve representational differences across videos.

    \item \textbf{We show how downstream MLPs shape the use of head outputs.} Ablation-induced representational changes correlate more strongly with head importance after MLP processing than immediately after the attention update. Fine-tuning MLPs alone partially restores performance when only lower-importance heads remain, suggesting improved use of their outputs.

    \item \textbf{We reveal collective contributions of local heads beyond individual importance rankings.} Across layers, local head sets support stronger perturbation responses at distant tokens than global head sets, and their joint ablation causes greater performance loss than ablating equally sized high-importance head sets. Fixed spatiotemporal kernels partially restore performance after ablation, indicating that stable routing supports part of their function.
\end{itemize}

\section{Related work}
\label{sec:related}

\noindent\textbf{Internal representations in vision and video Transformers.}
Prior work has used comparisons of representations across layers and attention analysis to reveal how information is organized within vision Transformers. Representations in image ViTs evolve with depth differently from those in convolutional networks~\cite{raghu2021vision}. Attention patterns and dense features in self-supervised vision Transformers reflect object regions, spatial structure, and semantic information, and are shaped by the pretraining objective~\cite{caron2021emerging,amir2022deep,park2023what,walmer2023teaching}. In video Transformers, Kowal \etal used concept discovery to reveal a progression from representations of spatiotemporal positions in early layers to object and event concepts in deeper layers~\cite{kowal2024understanding}. Layer-wise probes have also been used to assess how readily motion and physical properties can be decoded at different depths~\cite{joseph2026interpreting,musa2026what}. These studies characterize representational content and its distribution across layers, providing a foundation for analyzing the underlying internal computations.

\noindent\textbf{Specialization, importance, and redundancy of attention heads.}
Prior work has identified functional specialization among Transformer attention heads, with some heads exhibiting syntactic, coreferential, or positional attention patterns~\cite{clark2019bert,voita2019analyzing,kovaleva2019revealing}. Importance analysis and pruning methods have also been used to identify key components in vision Transformers~\cite{yu2023xpruner,palenzuela2025importance,li2023howdoes}. In video models, Kowal \etal combined concept discovery with joint masking to assess head importance and perform pruning~\cite{kowal2024understanding}; Joseph \etal used targeted head interventions to study physical and temporal reasoning abilities~\cite{joseph2026interpreting}. Functional relationships among heads have also received attention: individual head importance does not necessarily align with actual pruning outcomes~\cite{budhraja2020weak}, the functional roles of heads depend on other heads~\cite{nam2025causal}, and cooperation, interference, and redundancy have been observed among heads in vision Transformers~\cite{lu2026discovering,mcgrath2023hydra}. We focus on the gap between individual head importance and the collective functional contributions of head sets in video Transformers, testing whether the former can adequately predict the performance degradation caused by jointly ablating head sets.

\noindent\textbf{Information flow and cross-layer interactions in attention.}
Prior work has shown that understanding the computational roles of Transformer attention heads requires considering factors beyond attention weights, including residual connections, normalization, and feed-forward networks~\cite{jain2019attention,kobayashi2020attention,brunner2020identifiability}. Kobayashi \etal examined how residual connections and normalization affect representational updates, while Geva \etal revealed the roles of feed-forward networks in processing representations and forming predictions~\cite{kobayashi2021incorporating,geva2021transformer,geva2022transformer}. Related studies have also used attention rollout, cross-layer attribution, and computational circuit analysis to examine computational dependencies among tokens and network components across layers~\cite{abnar2020quantifying,chefer2021transformer,modarressi2022globenc,elhage2021mathematical}. In vision models, connections between local attention and convolution, along with spatiotemporal factorization and local window designs for video attention, further illustrate the roles of different interaction ranges and organizational patterns~\cite{cordonnier2020relationship,bertasius2021space,liu2022videoswin}. We combine head ablation, MLP interventions, and cross-layer perturbation analysis to study collective computation in video Transformers, examining how downstream processing shapes head contributions and how local interactions compose across layers to produce broader effects.

\section{Head importance}
\label{sec:functional-core}

In this section, we rank attention heads in video Transformers by importance using random joint ablation, providing a common basis for subsequent analyses. We examine whether important heads identified on a single task also affect other tasks, and how head sets with different levels of importance influence the structure of video representations. We further examine how well head importance can be predicted from computational properties.


\subsection{Importance estimation}

Functional redundancy among attention heads means that ablating a single head often produces only a small change in performance, making importance estimates sensitive to evaluation variability~\cite{michel2019sixteen,voita2019analyzing}. We therefore estimate the relationship between each head and task performance loss through repeated random joint ablation of different head sets.

We use the ViT-L encoders of V-JEPA, V-JEPA 2, VideoMAE, and VideoMAE V2, each comprising 24 Transformer blocks with 16 attention heads per block, for a total of 384 heads. During evaluation, we freeze the pretrained encoder and re-extract video representations under each ablation condition, then train a linear classifier to assess the linear decodability of task-relevant information after intervention~\cite{alain2018understanding,hewitt2019designing}.

We define four video understanding tasks covering general action and object recognition as well as fine-grained judgments of dynamics. \textbf{Action} evaluates action category recognition using a ten-class subset of Something-Something V2 (SSv2)~\cite{goyal2017something}. \textbf{Object} uses nine instrument-related activity classes from UCF101 to assess object- or scene-related semantic information~\cite{soomro2012ucf101}. \textbf{Direction} and \textbf{Outcome} are binary classification tasks based on SSv2, assessing the direction of object motion caused by an action (left or right) and whether the action is successfully completed, respectively. We estimate head importance separately on Action and Object to assess the consistency of the resulting rankings. All subsequent cross-task ablations and other analyses use the importance ranking derived from Action. Full dataset details and evaluation protocols are provided in Appendix~\ref{sec:app-tasks-ablation}.

In each trial, we randomly select 25\% of all attention heads for joint ablation, repeating the procedure for $N=3000$ trials. Each selected head's output is replaced with its mean over training samples and token positions to suppress sample- and position-specific information~\cite{zhang2024bestpractices}. Let $S_n$ denote the set of heads ablated in the $n$-th trial, and let $D(S_n)$ denote the drop in task accuracy relative to the unablated model (Intact). The importance $I(h)$ of head $h$ is defined as the average performance loss across all ablation trials that include that head:
\begin{equation}
I(h)=\frac{1}{N_h}\sum_{n=1}^{N}\mathbf{1}[h\in S_n]D(S_n),
\label{eq:head-importance}
\end{equation}
where $N_h$ is the number of trials in which head $h$ is included in the ablated set. We rank heads in descending order of $I(h)$ for subsequent cumulative ablation and functional analysis.


\subsection{Cross-task consistency}

\begin{figure}[!htbp]
    \centering
    \includegraphics[width=\columnwidth]{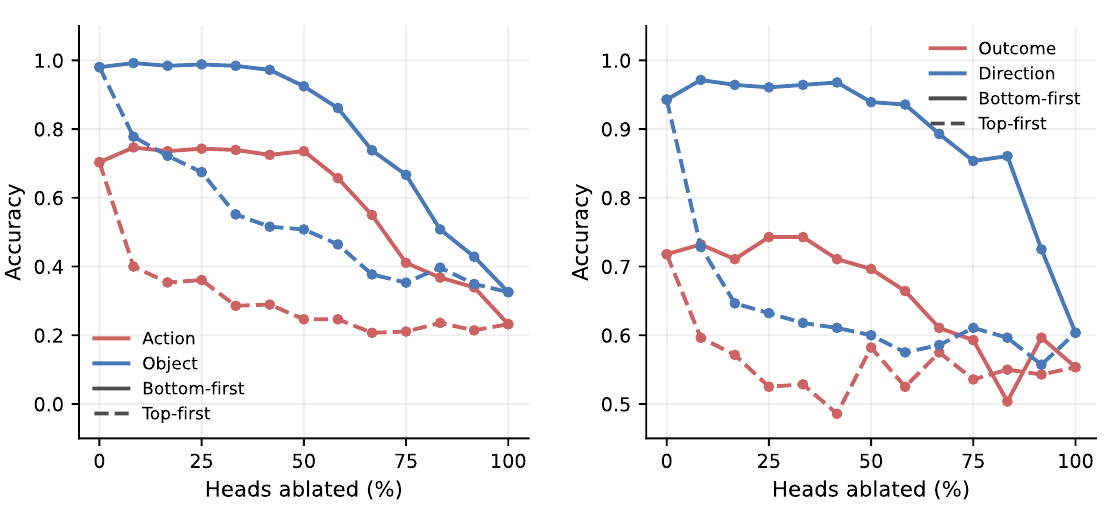}
    \caption{\textbf{Task performance under different attention head ablation orders.} Solid lines indicate cumulative ablation from lowest to highest importance (Bottom-first), while dashed lines indicate ablation from highest to lowest importance (Top-first).}
    \label{fig:cumulative}
\end{figure}

Using the importance ranking derived from Action, we examine how head ablation affects other video understanding tasks. As shown in Figure~\ref{fig:cumulative}, when low-importance heads are ablated first in V-JEPA, performance on all four tasks remains largely unchanged even after approximately half of the heads have been ablated. In contrast, ablating high-importance heads first causes accuracy to drop rapidly across all tasks. These results indicate that tasks relying on different types of video information nevertheless share a dependence on some key attention heads.

This pattern holds across all four video Transformers. Table~\ref{tab:action-rank-ablation-k96} shows that ablating the top 25\% of heads by importance also substantially reduces accuracy on the other tasks, whereas ablating the bottom 25\% generally has a smaller effect and even improves performance on some tasks. Comparing the importance rankings estimated independently on Action and Object, we find positive Spearman correlations in all four models, with 34--57 heads shared between their top 25\% sets, exceeding the overlap expected by chance in every model. These results indicate that multiple video understanding tasks rely on a shared subset of high-importance heads. Cumulative ablation results for the remaining models are provided in Figure~\ref{fig:app-cumulative} in the appendix.

\begin{table*}[!t]
  \centering
  \small

  \caption{\textbf{Cross-task ablation results for head sets of varying importance.} Values indicate accuracy changes relative to Intact (in percentage points, pp) after ablating the top or bottom 25\% of heads. The last two columns report the Spearman correlation between the Action and Object importance rankings and the overlap between their top 25\% head sets (expected overlap by chance: 24).}
  \label{tab:action-rank-ablation-k96}

  \setlength{\tabcolsep}{7pt}
  \renewcommand{\arraystretch}{1.08}
  \setlength{\heavyrulewidth}{1.2pt}
  \setlength{\lightrulewidth}{0.5pt}
  \setlength{\cmidrulewidth}{0.5pt}
  \setlength{\arrayrulewidth}{0.5pt}

  \newlength{\taskpad}
  \setlength{\taskpad}{4pt}

  \begin{tabular}{
    c
    >{\raggedright\arraybackslash}m{2.6cm}|
    >{\hspace{\taskpad}}c<{\hspace{\taskpad}}
    >{\hspace{\taskpad}}c<{\hspace{\taskpad}}
    >{\hspace{\taskpad}}c<{\hspace{\taskpad}}
    >{\hspace{\taskpad}}c<{\hspace{\taskpad}}|
    c c
  }
    \toprule
    \rule{0pt}{2.4ex}
    \textbf{Model} &
    \textbf{Ablated heads} &
    \textbf{Action} &
    \textbf{Object} &
    \textbf{Direction} &
    \textbf{Outcome} &
    $\boldsymbol{\rho}_{\textbf{Obj,Act}}$ &
    \textbf{Top 25\%} $\boldsymbol{\cap}$ \\
    \specialrule{0.5pt}{2pt}{1.5pt}
    \specialrule{0.5pt}{0pt}{2pt}

    \multirow{2}{*}{V-JEPA}
      & Top 25\%    & $-34.3$ & $-30.6$ & $-31.1$ & $-19.3$
      & \multirow{2}{*}{$0.61$} & \multirow{2}{*}{57} \\
      & Bottom 25\% & $+3.9$  & $+0.8$  & $+1.8$  & $+2.5$
      & & \\
    \midrule

    \multirow{2}{*}{VideoMAE}
      & Top 25\%    & $-25.7$ & $-51.6$ & $-10.0$ & $-11.8$
      & \multirow{2}{*}{$0.43$} & \multirow{2}{*}{46} \\
      & Bottom 25\% & $+6.1$  & $-5.2$  & $+2.5$  & $+2.1$
      & & \\
    \midrule

    \multirow{2}{*}{V-JEPA2}
      & Top 25\%    & $-27.9$ & $-26.6$ & $-20.7$ & $-16.1$
      & \multirow{2}{*}{$0.43$} & \multirow{2}{*}{51} \\
      & Bottom 25\% & $+12.9$ & $0.0$   & $+3.2$  & $+2.5$
      & & \\
    \midrule

    \multirow{2}{*}{VideoMAEv2}
      & Top 25\%    & $-23.9$ & $-11.1$ & $-8.9$  & $-6.1$
      & \multirow{2}{*}{$0.17$} & \multirow{2}{*}{34} \\
      & Bottom 25\% & $+8.6$  & $0.0$   & $+8.9$  & $+3.2$
      & & \\
    \bottomrule

  \end{tabular}
\end{table*}


\subsection{Representational structure}

\begin{figure}[!t]
    \centering
    \includegraphics[width=\columnwidth]{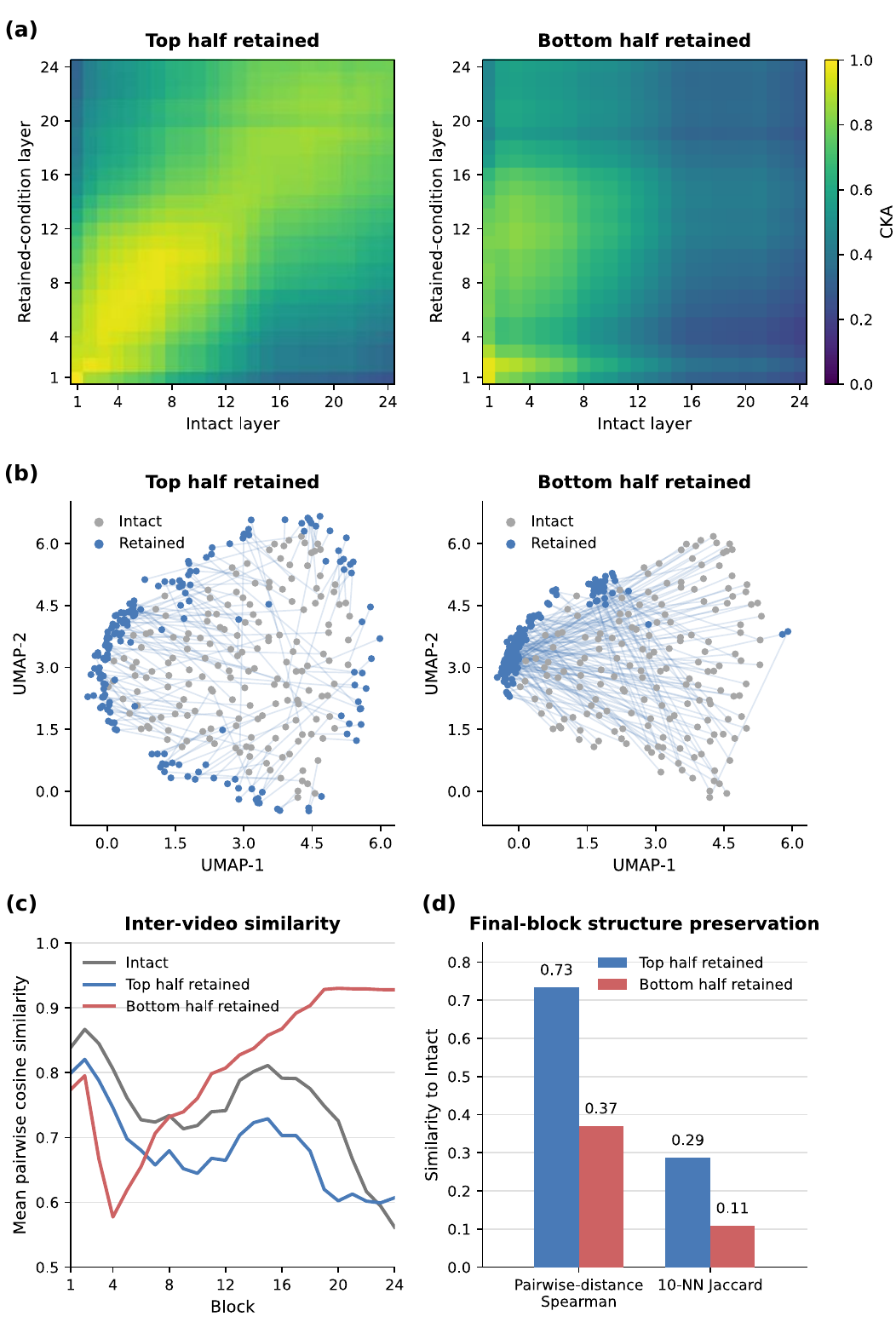}
    \caption{\textbf{Representational structure under head sets with different importance levels.}
    \textbf{(a)} Cross-layer CKA matrices between intervened and Intact representations.
    \textbf{(b)} Shared UMAP projections before and after ablation, using an Intact-fitted mapping.
    \textbf{(c)} Layer-wise mean pairwise cosine similarity across videos.
    \textbf{(d)} Spearman correlation of inter-video distance rankings and 10-NN Jaccard similarity, both relative to Intact.}
    \label{fig:geometry}
    \vspace{-40pt}
\end{figure}

To examine whether the cross-task roles of high-importance heads are associated with a shared internal representational structure, we retain only the top or bottom 50\% of attention heads by importance in separate conditions. We compare the resulting representations with those of Intact and assess how well the original structure of representational differences across videos is preserved.

High-importance head sets better preserve the original representations. As shown in Figure~\ref{fig:geometry}(a), when only high-importance heads are retained, representational similarity measured by centered kernel alignment (CKA)~\cite{kornblith2019similarity} between the intervened model and Intact remains high across most layers, particularly in deeper layers. Retaining only low-importance heads substantially reduces representational similarity in deeper layers. The shared UMAP~\cite{mcinnes2020umap} projections show a similar trend (Figure~\ref{fig:geometry}(b)): retaining high-importance heads yields a distribution of video samples closer to that of Intact, whereas retaining low-importance heads leads to larger sample displacements and a more concentrated distribution.

Figure~\ref{fig:geometry}(c) shows that retaining only low-importance heads substantially increases the mean pairwise cosine similarity between video representations in deeper layers, indicating that representations of different videos become more aligned in direction. Retaining high-importance heads better preserves representational differences across videos. Figure~\ref{fig:geometry}(d) further shows that, when high-importance heads are retained, both inter-video distance rankings and nearest-neighbor structure remain closer to those of Intact. These results indicate that high-importance head sets help maintain relative distances and local neighborhood relationships among video representations.

The other three video Transformers also show a relative advantage of high-importance head sets in preserving representational structure (Figure~\ref{fig:app-geometry} in the appendix). Together with the cross-task ablation results, these findings suggest that preserving representational distinctions across videos may be one reason why high-importance heads play important roles in multiple tasks.


\subsection{Importance prediction}

Building on the importance estimates above, we examine how the computational properties of attention heads relate to and predict their functional importance. We focus on V-JEPA, with results for the other models provided in Appendix~\ref{sec:cross-model-prediction}.

We extract ten attention head properties covering four aspects: spatiotemporal attention range, attention weight structure, outputs and representational responses, and network position. The complete list and definitions are provided in Appendix~\ref{sec:head-property-definitions}. Among these properties, mean spatial attention distance $d_{\mathrm{s}}$ and mean temporal attention distance $d_{\mathrm{t}}$ quantify the spatial and temporal extent of attention, respectively. $\Delta_{\mathrm{attn}}$ and $\Delta_{\mathrm{mlp}}$ measure the magnitudes of representational changes induced by the same head ablation, measured after the attention module and the MLP in the same block, respectively.

We use these properties to predict head importance $I(h)$ and evaluate generalization to unseen heads through repeated cross-validation. Predictive performance is measured by the Spearman rank correlation between predicted and actual importance rankings, and the contribution of each property is assessed through permutation importance~\cite{aaron2019allmodels}. Details of property selection, model training, and evaluation protocols are provided in Appendix~\ref{sec:app-prediction-protocol}.

\begin{figure}[!t]
    \centering
    \includegraphics[width=0.78\columnwidth]{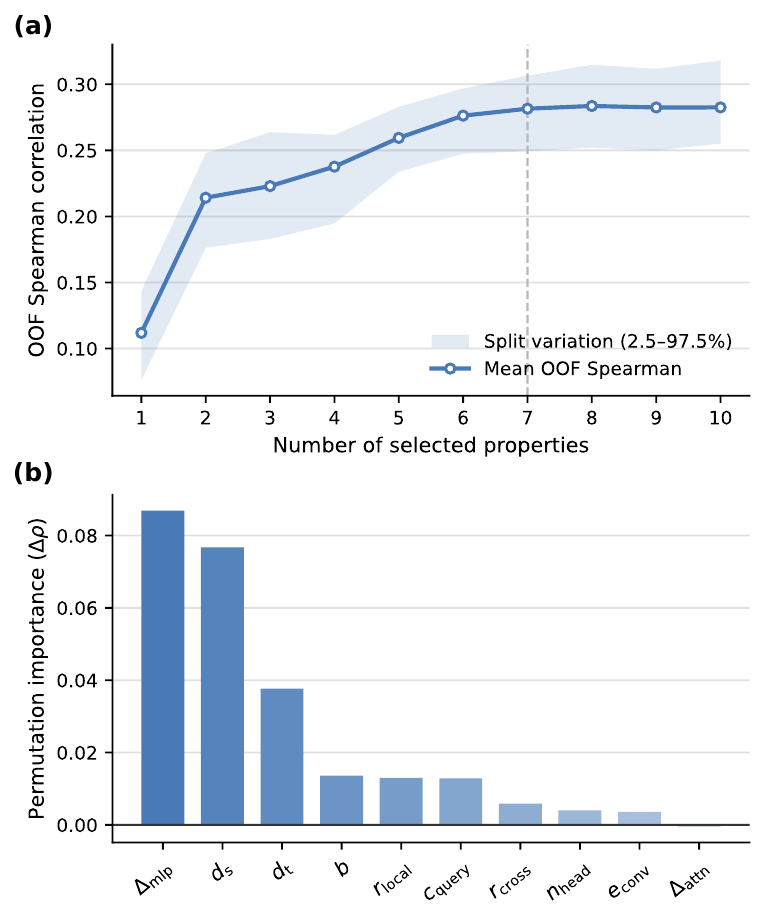}
    \caption{\textbf{Head importance prediction and property contributions.}
    \textbf{(a)} Spearman correlation between predicted and actual importance rankings for unseen heads, using different numbers of properties. Results are evaluated over 10 repetitions of six-fold cross-validation; intervals indicate variability across data splits.
    \textbf{(b)} Decrease in prediction correlation after randomly permuting each property individually when using seven properties.}
    \label{fig:properties}
\end{figure}

As shown in Figure~\ref{fig:properties}(a), predictive performance improves as more properties are included and begins to saturate at around seven properties. Figure~\ref{fig:properties}(b) shows that the post-MLP ablation-induced representational change $\Delta_{\mathrm{mlp}}$ and the mean spatial and temporal attention distances $d_{\mathrm{s}}$ and $d_{\mathrm{t}}$ provide the main predictive cues: heads with larger $\Delta_{\mathrm{mlp}}$ and shorter attention distances tend to be more important.
Results for the other video Transformers likewise identify spatial attention range as a predictive cue, while the contributions of other properties and overall predictive performance vary across models (Figure~\ref{fig:app-predictors} in the appendix).

Overall, prediction correlations remain modest, indicating that the computational properties of individual heads explain only part of the variation in head importance. The two types of predictive cues identified above warrant further investigation: post-MLP ablation-induced representational changes reflect how downstream computation processes head outputs, while spatiotemporal attention distances raise the question of how attention interactions with different ranges combine across layers. The next two sections examine these two directions in turn.

\section{Downstream computation}
\label{sec:downstream-computation}

\begin{figure*}[!t]
    \centering
    \includegraphics[width=0.78\textwidth]{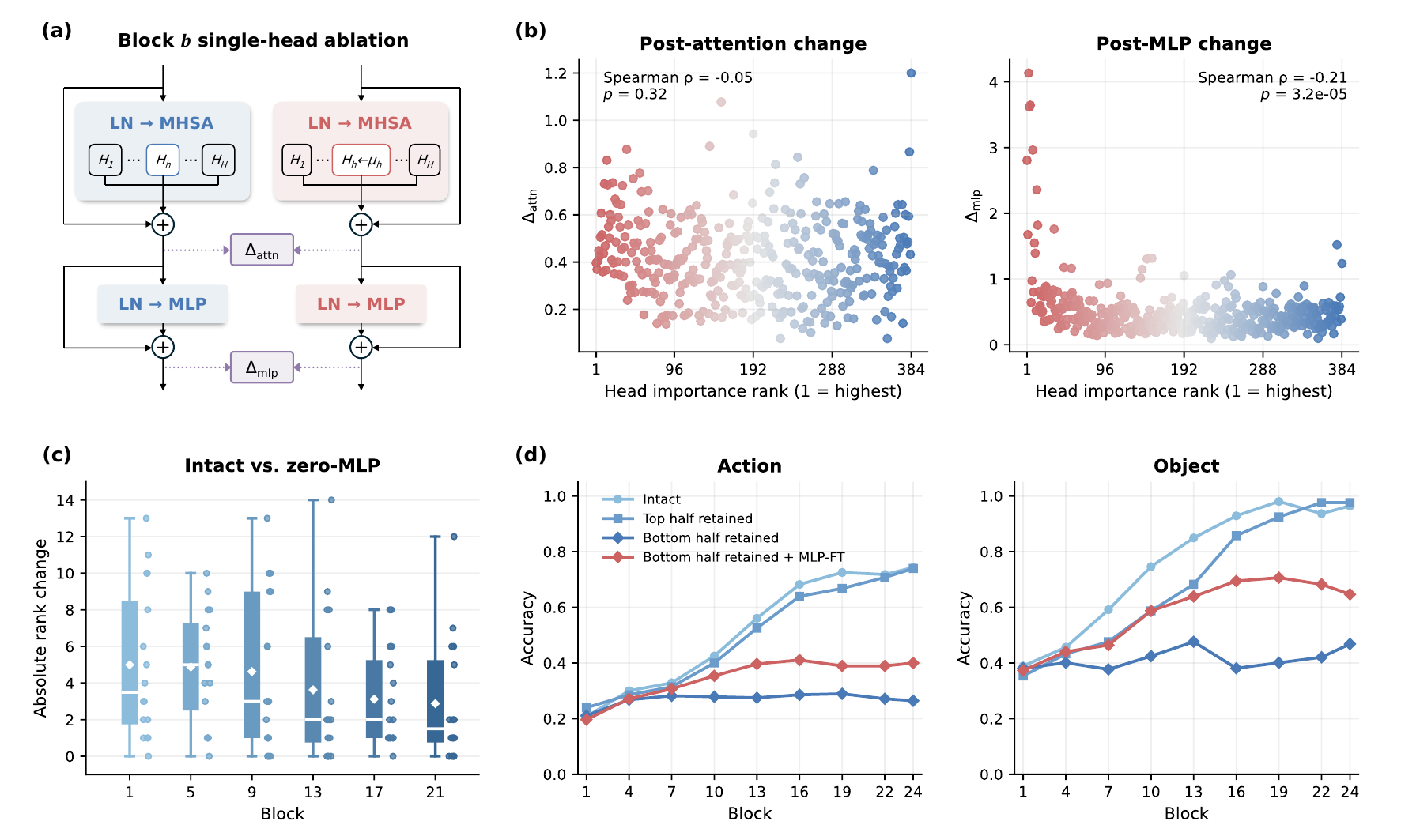}
    \caption{\textbf{Downstream MLP processing and the functional importance of attention heads.}
    \textbf{(a)} Measurement points for ablation-induced representational changes $\Delta_{\mathrm{attn}}$ and $\Delta_{\mathrm{mlp}}$.
    \textbf{(b)} Relationships between ablation-induced representational changes and head importance ranks. Each point represents a head. Spearman correlations $\rho$ and $p$-values are annotated.
    \textbf{(c)} Absolute changes in within-block head importance ranks after ablating the MLP in the same block, shown for six blocks. Box plots show the distributions, and white diamonds indicate the means.
    \textbf{(d)} Layer-wise accuracy on Action and Object under different head retention and MLP update conditions.}
    \label{fig:mlp}
\end{figure*}

Attention head outputs are processed through residual connections, MLPs, and subsequent blocks, so their functional importance may depend on how the downstream network uses these signals. The post-MLP ablation-induced representational change $\Delta_{\mathrm{mlp}}$ is one of the main predictive cues for head importance (Figure~\ref{fig:properties}(b)). In this section, we compare ablation-induced representational changes before and after the MLP and combine MLP interventions with self-supervised adaptation experiments to examine how downstream computation shapes the functional contributions of attention heads.

As shown in Figure~\ref{fig:mlp}(a), we compare the representational changes induced by the same head ablation after the attention residual addition and after the MLP residual addition in the same block, denoted by $\Delta_{\mathrm{attn}}$ and $\Delta_{\mathrm{mlp}}$, respectively. Figure~\ref{fig:mlp}(b) shows that the Spearman correlation between $\Delta_{\mathrm{attn}}$ and head importance rank is only $-0.05$, compared with $-0.21$ for $\Delta_{\mathrm{mlp}}$, with a few high-importance heads exhibiting exceptionally large post-MLP representational changes. These results indicate that representational changes after MLP processing better reflect the functional characteristics of high-importance heads than the immediate changes following the attention update.

We set the output of a Transformer block's MLP branch to zero, so that representations pass through the MLP sublayer solely via the residual connection, and re-estimate the importance ranking of heads within that block. Figure~\ref{fig:mlp}(c) shows substantial changes in the within-block ranks of some heads across the six selected blocks, indicating that the computational conditions following the attention update influence the relative functional importance of attention heads.

With only the bottom half of attention heads by importance retained, we fine-tune only the MLP parameters using V-JEPA's original feature prediction pretraining objective, keeping all other encoder parameters frozen (see Appendix~\ref{sec:app-mlp-finetuning} for training details). Self-supervised adaptation partially restores performance on Action and Object using features from each layer (Figure~\ref{fig:mlp}(d)). This indicates that, after high-importance heads are ablated, downstream MLPs can still adapt through self-supervision to improve the usability of task-relevant information in the remaining representations.

\section{Local attention across layers}
\label{sec:local-transport}

Shorter spatial and temporal attention distances are associated with greater head importance (Figure~\ref{fig:properties}(b)), raising the question of how local interactions support video understanding through composition across layers. In this section, we examine the collective computation of local attention heads through long-range perturbation responses, joint ablation, and fixed attention routing.

\subsection{Long-range interactions}

\begin{figure*}[!t]
    \centering
    \includegraphics[width=\textwidth]{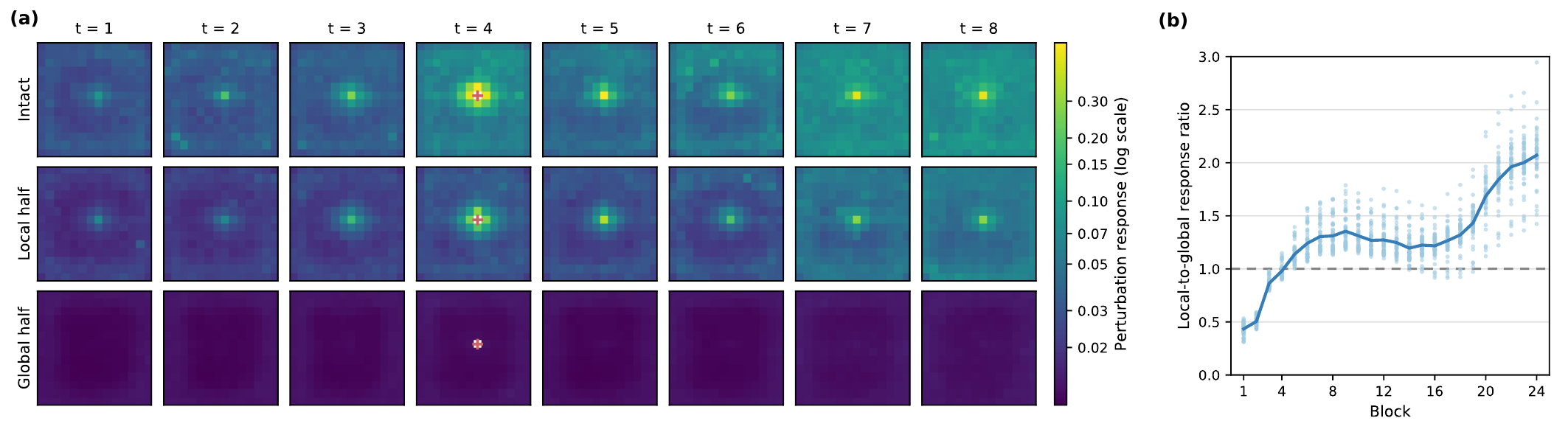}
    \caption{\textbf{Perturbation responses under different head sets.}
    \textbf{(a)} Final-layer response distributions under three conditions. Red crosses mark the perturbation source, whose position is left blank.
    \textbf{(b)} Layer-wise ratios of Local half to Global half response magnitudes at the 52 boundary tokens in the temporal slice containing the perturbation source. Scatter points show per-token ratios, and the line shows their mean.}
    \label{fig:diffusion}
\end{figure*}

The effective spatial extent of network computation can emerge through the composition of local operations across layers, as previously studied in convolutional neural networks~\cite{luo2016understanding}. We first examine how local attention heads support interactions between distant tokens across layers. We rank all attention heads by mean spatial attention distance $d_{\mathrm{s}}$ and retain either the half with the shortest distances or the half with the longest distances, denoted as Local half and Global half, respectively. We apply a small perturbation to the central token of the input sequence and measure the representational responses of other tokens at different network depths to assess the range and strength of interactions across positions supported by the two
head sets. Response magnitudes are computed using central finite differences; the detailed protocol is provided in Appendix~\ref{sec:app-perturbation-protocol}.

Despite their shorter single-step attention range, local head sets can support stronger interactions between distant tokens through computation across multiple layers. Figure~\ref{fig:diffusion}(a) shows the response distributions in the final block under three conditions: responses under Local half extend from the vicinity of the perturbation source to distant spatial regions and other time slices, whereas Global half exhibits weaker responses overall. To quantitatively compare the long-range responses supported by the two head sets, we compute the ratio of Local half to Global half response magnitudes at the 52 tokens along the outer boundary of the $14\times14$ grid in the time slice containing the perturbation source (see Appendix~\ref{sec:app-perturbation-protocol} for calculation details). Figure~\ref{fig:diffusion}(b) shows the per-token ratios and their mean. In early layers, Global half produces stronger long-range perturbation responses than Local half, consistent with its broader single-step spatial attention range. As network depth increases, local interactions combine across layers to propagate the perturbation's effects to more distant positions, allowing the long-range responses under Local half to gradually surpass those under Global half. These results show that short-range interactions can produce widespread and stronger long-range effects through composition across layers.

\subsection{Collective contributions}

\begin{figure*}[!t]
    \centering
    \includegraphics[width=0.83\textwidth]{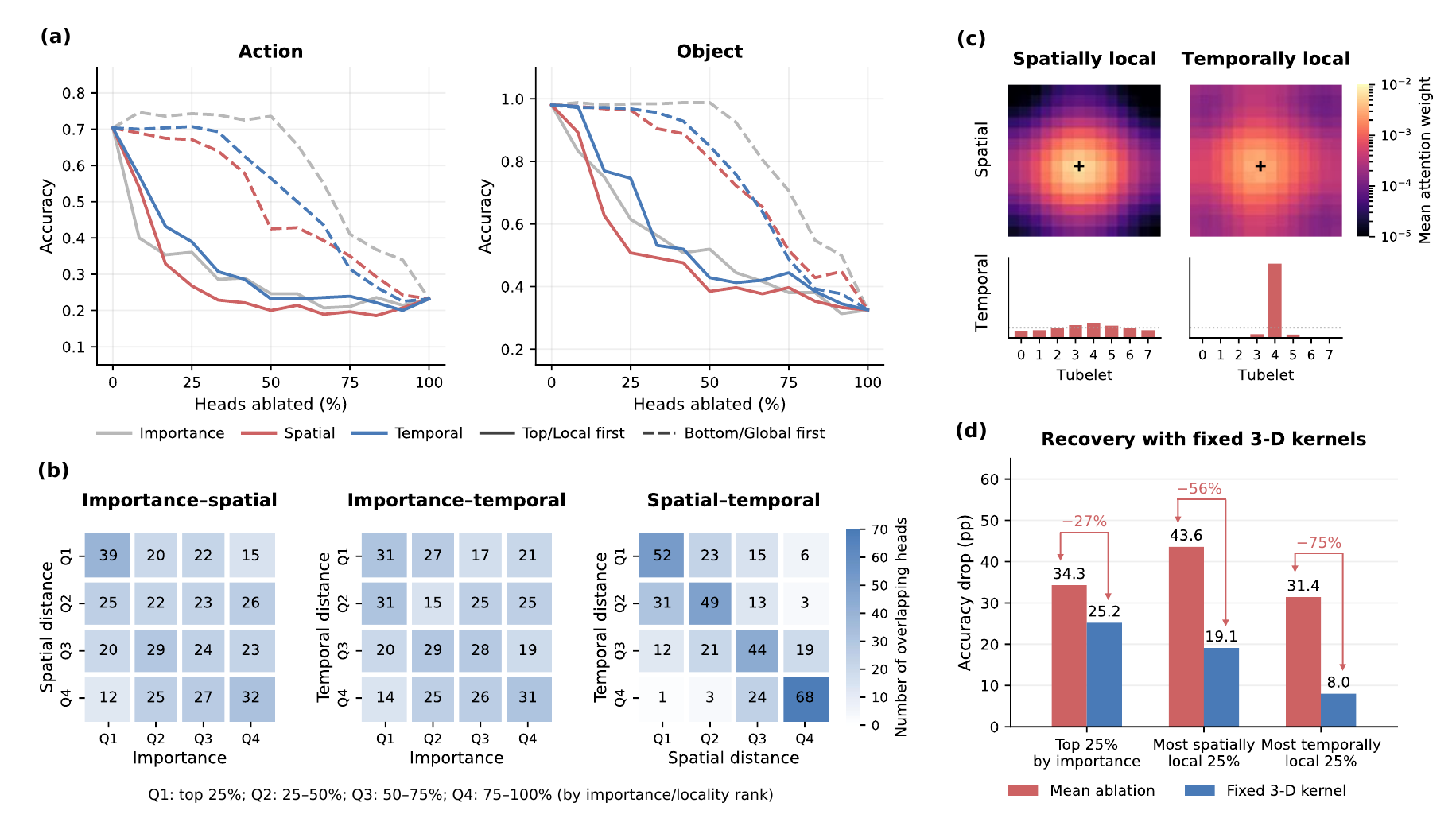}
    \caption{\textbf{Performance loss from ablating head sets and recovery with fixed routing.}
    \textbf{(a)} Cumulative ablation results with heads ranked by importance, spatial attention distance, or temporal attention distance.
    \textbf{(b)} Cross-tabulations of quartile groups for each pair of metrics. Q1 through Q4 correspond to importance from highest to lowest and attention distance from shortest to longest. Numbers indicate head counts in the corresponding groups.
    \textbf{(c)} Two example attention distributions, one for a spatially local head and one for a temporally local head. Crosses mark query positions; heatmaps and bar plots show the query token's mean spatial and temporal attention distributions, respectively.
    \textbf{(d)} Action accuracy loss (pp) under mean-replacement ablation and replacement with fixed spatiotemporal kernels for three head sets: the top 25\% by importance, the most spatially local 25\%, and the most temporally local 25\%. Arrows indicate the fraction of the ablation-induced loss recovered.}
    \label{fig:collective}
\end{figure*}

To examine whether the cross-layer role of local heads is reflected in downstream task performance, we compare performance degradation under cumulative ablation of different head sets. As shown in Figure~\ref{fig:collective}(a), ablating spatially or temporally local heads first causes substantial performance degradation on both main tasks. As the ablated fraction increases, accuracy drops particularly sharply when spatially local heads are ablated first, with the curve even falling below that obtained by ablating heads in descending order of importance. These results indicate that individual head importance rankings cannot fully predict the performance degradation caused by jointly ablating head sets and may underestimate the overall contribution of spatially local head sets to task performance. Ablating global heads first still causes greater performance loss than ablating low-importance heads first, indicating that global heads also make functional contributions.

Figure~\ref{fig:collective}(b) shows limited correspondence between head importance quartiles and spatial or temporal attention distance quartiles, indicating that the substantial performance loss from ablating local head sets cannot be readily explained by their overlap with high-importance heads. Spatial and temporal attention distances are more closely aligned, with heads concentrated mainly in the same or adjacent quartiles across the two dimensions. This also provides insight into the responses across time slices in Figure~\ref{fig:diffusion}(a): although Local half is selected by spatial attention distance and also tends to have a short temporal attention range, perturbation effects can still extend to other time slices through local interactions across multiple layers. Some tasks in the other three models are also relatively sensitive to the ablation of local head sets (Figure~\ref{fig:app-routing-ablation} in the appendix). These results further indicate that individual head importance does not fully align with collective functional contributions, and that the role of local heads must be understood in terms of their composition across layers.

\subsection{Fixed attention routing}

Jointly ablating local head sets causes substantial performance loss, but do their functions depend on input-dependent attention routing~\cite{tay2021synthesizer,liu2021payattention}? Figure~\ref{fig:collective}(c) shows two examples, one spatially local head and one temporally local head: the query token's mean attention is concentrated in its spatial neighborhood or at nearby temporal positions, exhibiting local aggregation patterns resembling convolutional kernels. This suggests that relatively stable spatiotemporal routing may support some of the functions of local heads.

To test this hypothesis, we restore attention computation in the ablated heads using fixed three-dimensional spatiotemporal kernels. We first compute each head's mean attention matrix, then average the weights over query-key position pairs with the same relative spatiotemporal displacement to construct fixed kernels, followed by row-wise normalization. After replacement, the attention weights no longer vary with the video input, but they still aggregate the current video's value vectors through weighted summation, preserving input-dependent content information.

Figure~\ref{fig:collective}(d) shows that ablating the most spatially local 25\% of attention heads causes a greater loss in task accuracy than ablating the top 25\% by importance (43.6 vs.\ 34.3 percentage points). Restoring computation with fixed spatiotemporal kernels recovers 56\% of the loss for the former set, substantially more than the 27\% recovered for the latter. For the most temporally local 25\% of heads, fixed kernels recover 75\% of the ablation-induced loss. These results indicate that a substantial portion of the functions of local head sets can be supported by stable routing for local information aggregation, without relying entirely on input-dependent attention weights.
\section{Conclusion, limitations and future work}
\label{sec:conclusion}

This work investigates collective computation in video Transformers by examining how head outputs are used downstream and how interactions compose across layers. Across four models, high-importance heads support multiple tasks and preserve representational distinctions across videos. Further analyses of V-JEPA show that MLP processing shapes head importance, while MLP-only fine-tuning partially restores performance when lower-importance heads remain. Local interactions compose across layers to support long-range perturbation responses and collective contributions beyond individual importance rankings. Partial recovery with fixed spatiotemporal kernels further suggests that stable routing supports some functions of local heads. Our study has two limitations: evaluations cover limited task subsets and use linear probes, while ablation and perturbation responses reveal functional dependencies without fully identifying the computational pathways. Future work could trace how attention heads and MLPs jointly transform local spatiotemporal information into object, motion, and event representations, and examine how these pathways support temporal reasoning across broader video tasks.

\section*{Acknowledgments}
{\raggedright
This study is supported by the National Natural Science
Foundation of China (52572413).\par
}
{
    \small
    \bibliographystyle{ieeenat_fullname}
    \bibliography{main}
}

\clearpage
\hypersetup{pageanchor=false}
\setcounter{page}{1}
\maketitlesupplementary
\par\small
\appendix

\section{Experimental protocols}
\label{sec:app-protocol}

\subsection{Evaluation tasks and ablation protocol}
\label{sec:app-tasks-ablation}

We define four tasks: Action, Object, Direction, and Outcome.
Action uses ten Something-Something V2 classes covering object motion, deformation, and manipulation (e.g., throwing, folding, and pouring), while Object uses nine instrument-related UCF101 activity classes.
For both tasks, we sample 100 videos per class.
Direction and Outcome are binary tasks constructed by merging original Something-Something V2 classes.
Direction groups four classes involving pushing or pulling objects to the left or right according to motion direction.
Outcome groups five types of actions (picking up, pouring out, attaching, tearing, and opening) and their corresponding pretend or failed-action classes into two categories: successful and unsuccessful execution.
Both binary tasks contain 500 videos per class.
All tasks are class-balanced, with videos split into training and test sets at an 18:7 ratio.

For each video, we average-pool the encoder's output token representations to obtain a video-level feature.
Our linear probe combines feature standardization with a linear classifier trained using SGD.
We refit the probe under each intervention condition and evaluate accuracy on a fixed test set.
Attention head ablation uses mean replacement: for each head, we compute its mean output over all training videos and token positions in the Intact model, then replace the ablated head's output with this mean to suppress its sample- and position-specific information.
For both Action and Object, we perform 3000 random joint ablation trials on each of the four models.
In each trial, we randomly ablate 25\% of the attention heads and record the accuracy drop relative to Intact.

\subsection{Predicting and evaluating head importance}
\label{sec:app-prediction-protocol}

For each model, we predict attention heads' percentile importance ranks on Action using repeated six-fold cross-validation to evaluate generalization to unseen heads.
Within each training fold, we select properties using maximum relevance minimum redundancy (mRMR)~\cite{peng2005feature} and fit an ExtraTrees regressor~\cite{breiman2001random,geurts2006extremely} with 200 extremely randomized trees.
Selection and fitting use only training-fold data; evaluation uses Spearman correlation between predicted and ground-truth importance ranks on the test fold.

Figure~\ref{fig:properties}(a) evaluates prediction using one to ten properties over 10 cross-validation repetitions~\cite{aaron2019allmodels}.
Mean Spearman correlation rises from 0.112 with one property to 0.282 with seven, then plateaus at 0.284 and 0.282 with eight and ten, respectively.
We therefore use seven properties for permutation importance analysis.

Figure~\ref{fig:properties}(b) assesses property contributions over 50 six-fold cross-validation repetitions, selecting seven properties per training fold.
Each selected property is permuted ten times within the test fold; its contribution is the mean decrease in predictive Spearman correlation.
Unselected properties receive zero importance, and results are aggregated across all ten candidates.
Mean predictive correlation is 0.279, with a variation interval of $[0.249, 0.307]$ across data splits.

\subsection{Self-supervised MLP fine-tuning}
\label{sec:app-mlp-finetuning}

The self-supervised fine-tuning of the Bottom half retained model in Figure~\ref{fig:mlp}(d) uses the full SSv2 training set without action or object labels, retaining V-JEPA's masked feature prediction objective, predictor, and exponential moving average (EMA) target encoder.
We update only the MLP parameters in the context encoder and predictor, freezing all other parameters, including those of the attention modules, patch embedding, positional encodings, and normalization layers.
Inputs are 16-frame clips cropped to $224\times224$, with masking and data augmentation following the original pretraining pipeline.
We train for one epoch (2639 iterations) with a batch size of 64, warming up the learning rate to $1.5\times10^{-4}$ over 0.1 epoch and then decaying it to $10^{-6}$.

\subsection{Perturbation responses at distant tokens}
\label{sec:app-perturbation-protocol}

For Figure~\ref{fig:diffusion}, we use a subset of 174 videos from Something-Something V2, with one video per class.
We rank all 384 attention heads by their mean spatial attention distance $d_{\mathrm{s}}$ and retain either the half with the shortest distances (Local half) or the half with the longest distances (Global half), comparing both with the unablated model (Intact).

For each sampled view $u$, let $\mathbf z_u$ denote the input token sequence.
We apply a small perturbation at the central position $s$, with magnitude
$\epsilon_u=\lambda\|\mathbf z_u(s)\|_2$,
where $\lambda=0.02$, so that the perturbation norm is $2\%$ of the source token's representation norm.
We apply positive and negative perturbations along each of four unit directions and use central finite differences to measure the response at each output position.
Let $F_{\ell,u,k}^{(c,\pm)}(i)$ denote the output representation at position $i$ of ViT block $\ell$ under condition $c$ after applying a positive or negative perturbation along direction $k$, where
$c\in\{\mathrm{Intact},\mathrm{Local},\mathrm{Global}\}$.
Averaging over the four perturbation directions and all sampled views gives the response magnitude:
\begin{equation}
g_\ell^{(c)}(i)=
\frac{1}{4|\mathcal U|}
\sum_{u\in\mathcal U}\sum_{k=1}^{4}
\frac{
\left\|
F_{\ell,u,k}^{(c,+)}(i)
-
F_{\ell,u,k}^{(c,-)}(i)
\right\|_2
}{2\epsilon_u},
\end{equation}
Here, $\mathcal U$ denotes the set of all sampled views.
This metric measures the mean representational response magnitude at each output position per unit input perturbation.
To quantitatively compare responses at spatially distant positions, we focus on the time slice containing the perturbation source, $t=3$, and define $\mathcal B$ as the set of 52 distinct tokens on the outer boundary of its $14\times14$ spatial grid.
We first compute the ratio of the mean response magnitude for Local half to that for Global half at each position, then average these ratios over all boundary positions:
\begin{align}
r_\ell(i)
&=
\frac{g_\ell^{(\mathrm{Local})}(i)}
{g_\ell^{(\mathrm{Global})}(i)},\\
R_\ell
&=
\frac{1}{|\mathcal B|}
\sum_{i\in\mathcal B}r_\ell(i),
\qquad |\mathcal B|=52.
\end{align}

\section{Head property definitions}
\label{sec:head-property-definitions}

Drawing on prior work, we define ten attention head properties across four categories: spatiotemporal attention range, attention weight structure, head outputs and representational responses, and network position.
These properties are summarized in Table~\ref{tab:head-properties}.
In what follows, $h$ denotes an attention head, $x$ an input video, and $i$ and $j$ the query and key positions, respectively.
$A_{ij}^{(h,x)}$ denotes the softmax-normalized attention weight.
$Q_x$ is the set of query positions for video $x$, and $K_i$ is the set of key positions accessible to query position $i$.

\subsection{Spatiotemporal attention range}
\label{sec:app-attention-range}

We first quantify the spatiotemporal attention range of each head using attention-weighted mean distances.
Let $\mathbf p_i$ denote the pixel coordinates of the center of the image patch corresponding to token $i$, and let $t_i$ denote its tubelet index.
The mean spatial and temporal attention distances are defined, respectively, as
\begin{align}
d_{\mathrm{s}}^{(h)}
&=\mathbb E_x\frac{1}{|Q_x|}\sum_{i\in Q_x}
\sum_{j\in K_i}A_{ij}^{(h,x)}
\lVert\mathbf p_j-\mathbf p_i\rVert_2,\\
d_{\mathrm{t}}^{(h)}
&=\mathbb E_x\frac{1}{|Q_x|}\sum_{i\in Q_x}
\sum_{j\in K_i}A_{ij}^{(h,x)}|t_j-t_i|.
\end{align}

We also examine the distribution of attention weights across temporal positions and within local neighborhoods in the same temporal slice.
The cross-temporal attention ratio $r_{\mathrm{cross}}$ measures the fraction of attention weight assigned to temporal slices other than the query's own and is defined as
\begin{equation}
r_{\mathrm{cross}}^{(h)}
=\mathbb E_x\frac{1}{|Q_x|}\sum_{i\in Q_x}
\sum_{j\in K_i}A_{ij}^{(h,x)}\mathds{1}[t_j\neq t_i].
\end{equation}
The same-slice local attention ratio $r_{\mathrm{local}}$ measures the fraction of attention weight assigned to the local spatial neighborhood within the query's temporal slice and is defined as
\begin{equation}
r_{\mathrm{local}}^{(h)}
=\mathbb E_x\frac{1}{|Q_x|}\sum_{i\in Q_x}
\sum_{j\in K_i}A_{ij}^{(h,x)}
\mathds{1}[t_j=t_i]\mathds{1}[\lVert\mathbf q_j-\mathbf q_i\rVert_2\leq1],
\end{equation}
Here, $\mathbf q_i$ denotes the integer patch-grid coordinates of token $i$.
The local neighborhood is defined by a Euclidean radius of one grid unit, including the current position and its four immediate horizontal and vertical neighbors, but excluding diagonal neighbors.

\subsection{Attention weight structure}
\label{sec:app-attention-structure}

We use the convolutional approximation error $e_{\mathrm{conv}}$ to quantify how closely a head's mean attention weight distribution matches a spatiotemporal kernel shared across positions.
First, we compute each head's mean attention matrix over the training set, $\bar A^{(h)}=\mathbb E_x[A^{(h,x)}]$.
For each relative spatiotemporal offset $\delta=(\Delta t,\Delta y,\Delta x)$, we define the shared kernel $K_h(\delta)$ by averaging the attention weights over all valid query--key position pairs with that offset.
We then reconstruct the attention matrix $A_{\mathrm{conv}}^{(h)}$ using zero padding and row-wise normalization, and define
\begin{equation}
e_{\mathrm{conv}}^{(h)}
=\frac{\lVert A_{\mathrm{conv}}^{(h)}-\bar A^{(h)}\rVert_F}
{\lVert\bar A^{(h)}\rVert_F}.
\end{equation}
A smaller error indicates that the head's mean attention distribution more closely resembles a 3D convolutional routing pattern shared across positions.

Query pattern consistency $c_{\mathrm{query}}$ measures the consistency of a head's relative attention distributions across query positions.
For each query position $i$, we aggregate and normalize its attention weights by relative spatiotemporal offset $\delta$ to obtain a distribution $P_i$.
We then compute the mean distribution across all query positions within each video, $\bar P=|Q_x|^{-1}\sum_{i\in Q_x}P_i$.
The metric is defined as
\begin{equation}
c_{\mathrm{query}}^{(h)}
=1-\mathbb E_x\frac{1}{|Q_x|}\sum_{i\in Q_x}
\operatorname{JSD}_2(P_i,\bar P),
\end{equation}
Here, $\operatorname{JSD}_2$ denotes the Jensen--Shannon divergence computed using base-2 logarithms.
Thus, $c_{\mathrm{query}}\in[0,1]$, with larger values indicating greater consistency of relative attention patterns across query positions.

\subsection{Outputs and representational responses}
\label{sec:app-representational-responses}

We quantify the effects of attention heads at different computational stages through the representational changes induced by head ablation.
Let $R_{\ell,i}^{\mathrm{attn}}(x)$ and $R_{\ell,i}^{\mathrm{mlp}}(x)$ denote the token representations at position $i$ for video $x$ after the attention and MLP modules of Transformer block $\ell$, respectively.
A tilde denotes the corresponding representation after ablating head $h$.
The two ablation-induced representational responses are defined as
\begin{align}
\Delta_{\mathrm{attn}}^{(h)}
&=\mathbb E_x\frac{1}{|Q_x|}
\sum_{i\in Q_x}
\left\|
R_{\ell,i}^{\mathrm{attn}}(x)
-\widetilde R_{\ell,h,i}^{\mathrm{attn}}(x)
\right\|_2,\\
\Delta_{\mathrm{mlp}}^{(h)}
&=\mathbb E_x\frac{1}{|Q_x|}
\sum_{i\in Q_x}
\left\|
R_{\ell,i}^{\mathrm{mlp}}(x)
-\widetilde R_{\ell,h,i}^{\mathrm{mlp}}(x)
\right\|_2.
\end{align}
Both metrics use the same mean-replacement head ablation, measuring representational changes after the attention module and after the MLP module in the same block, respectively.

We also use the projected output norm $n_{\mathrm{head}}$ to measure the magnitude of each attention head's output contribution.
Let $O_{h,i}(x)$ denote the output of head $h$ at position $i$ before head concatenation, and let $W_O^{(h)}$ denote the corresponding weight block in the multi-head attention output projection matrix.
The head's projected output at position $i$ is $z_{h,i}^{(x)}=O_{h,i}(x)W_O^{(h)}$, with the shared output bias excluded from individual head contributions.
We define
\begin{equation}
n_{\mathrm{head}}^{(h)}
=\mathbb E_x\frac{1}{|Q_x|}
\sum_{i\in Q_x}
\left\|z_{h,i}^{(x)}\right\|_2.
\end{equation}

\subsection{Network position}
\label{sec:app-network-position}

We use normalized depth to represent an attention head's position within the network.
For a model with $L=24$ Transformer blocks, the normalized depth of a head in block $\ell$ is defined as
\begin{equation}
b^{(h)}=\frac{\ell}{L},
\end{equation}
where $\ell\in\{1,\ldots,L\}$.

In the importance prediction experiments, all nine properties other than normalized depth $b$ are converted to percentile ranks across all 384 attention heads within each model; $b$ retains its original definition.

\section{Cross-model results}
\label{sec:cross-model}

\subsection{Head importance}
\label{sec:cross-model-importance}

\begin{figure*}[p]
  \centering
  \includegraphics[width=\textwidth]{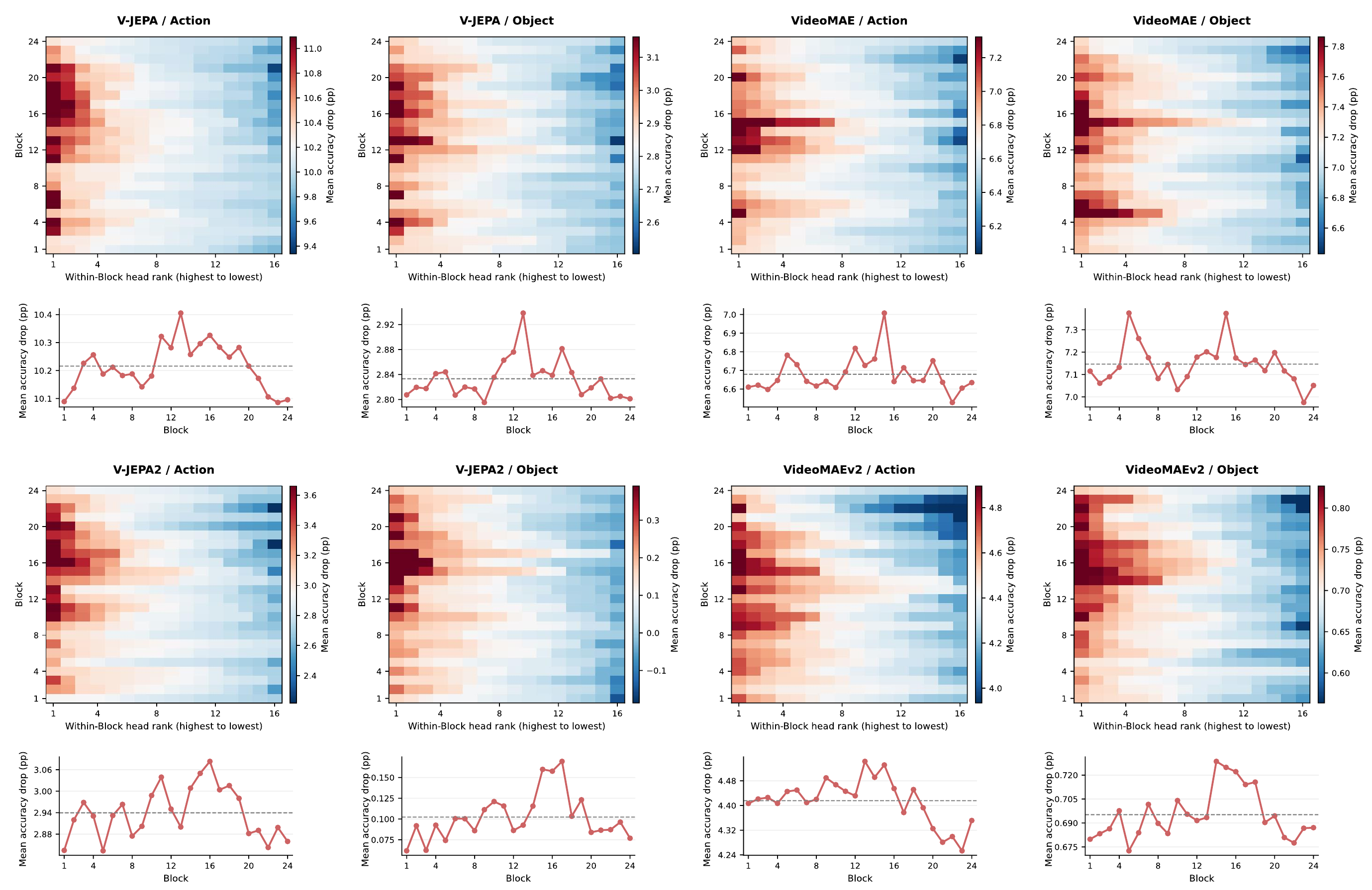}
  \caption{\textbf{Head importance distributions on Action and Object across four models.} Importance is measured as the mean accuracy drop across random joint ablation trials containing the head. In each heatmap, heads within each block are sorted by decreasing importance. The curves below show the mean importance within each block.}
  \label{fig:app-importance-maps}
\end{figure*}

Figure~\ref{fig:app-importance-maps} shows the distribution of attention head importance on Action and Object across the four video Transformers.
All models exhibit pronounced variation both within and across layers: heads in the same block differ in importance, while the mean importance within each block also varies with network depth.
High-importance heads tend to occur more frequently in the middle or middle-to-late layers, although their distribution varies across models and tasks.
For example, in VideoMAE, mean head importance peaks in the middle layers on Action but in shallower layers on Object.
These results indicate that attention head importance exhibits structure across layers, alongside variation within layers that cannot be captured by network depth alone.

\begin{figure*}[p]
  \centering
  \includegraphics[width=0.65\textwidth]{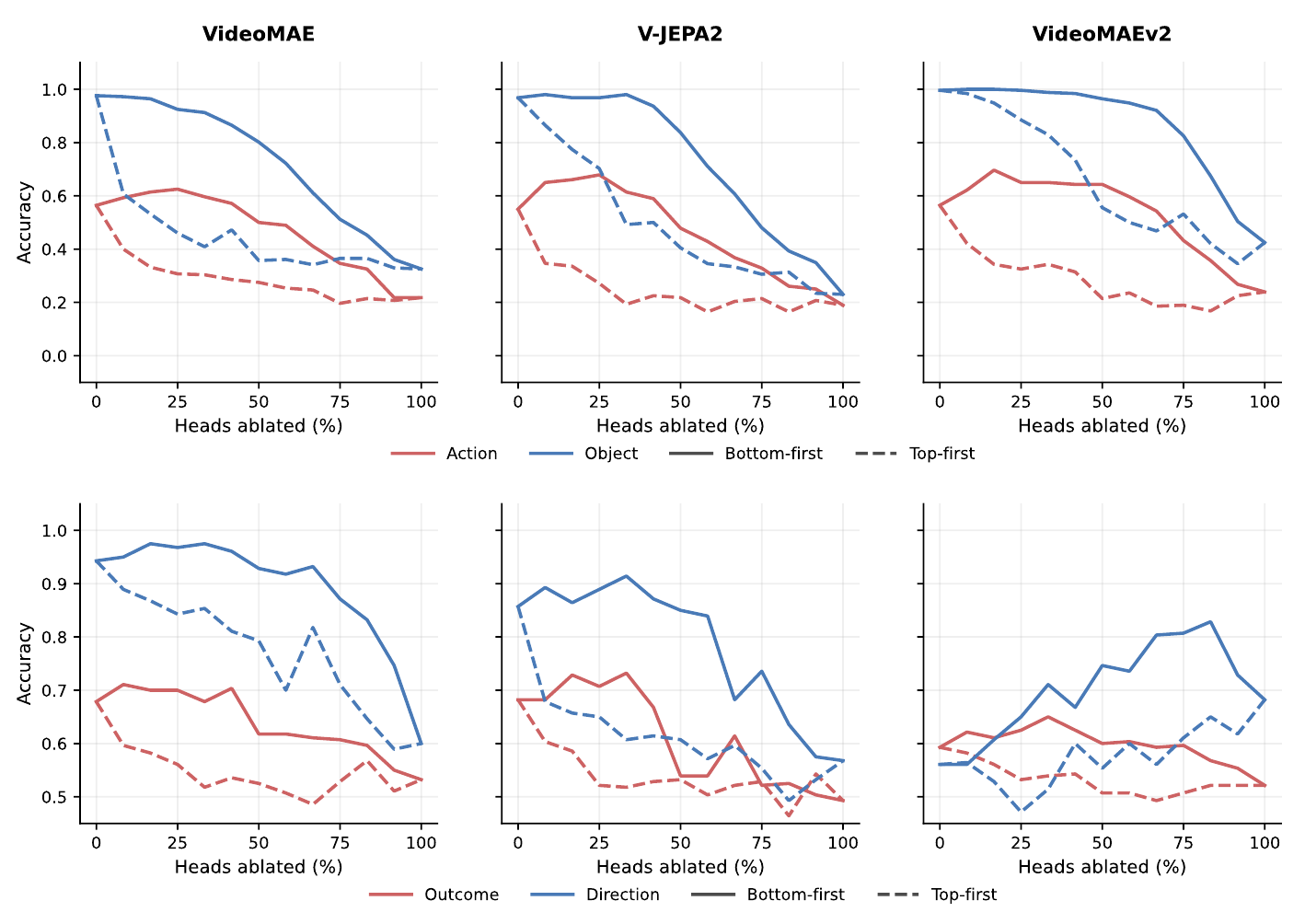}
  \caption{\textbf{Cross-task cumulative ablation results across models.} We perform cumulative ablation using the importance ranking derived from Action, ablating heads from highest to lowest importance (Top-first) or lowest to highest importance (Bottom-first).}
  \label{fig:app-cumulative}
\end{figure*}

Figure~\ref{fig:app-cumulative} provides additional cross-task cumulative ablation results for VideoMAE, V-JEPA 2, and VideoMAE V2.
All curves use the importance ranking derived from Action, cumulatively ablating heads from either highest to lowest or lowest to highest importance.
Ablating high-importance heads first causes accuracy to decline rapidly on multiple tasks.
In contrast, ablating low-importance heads first generally leads to a slower decline and can even improve performance on some tasks.
This trend is consistent with the V-JEPA results in the main text, indicating that high-importance heads identified on a single task also play important roles in other video understanding tasks.

\subsection{Representational structure}
\label{sec:cross-model-representations}

\begin{figure*}[p]
  \centering
  \includegraphics[width=0.9\textwidth]{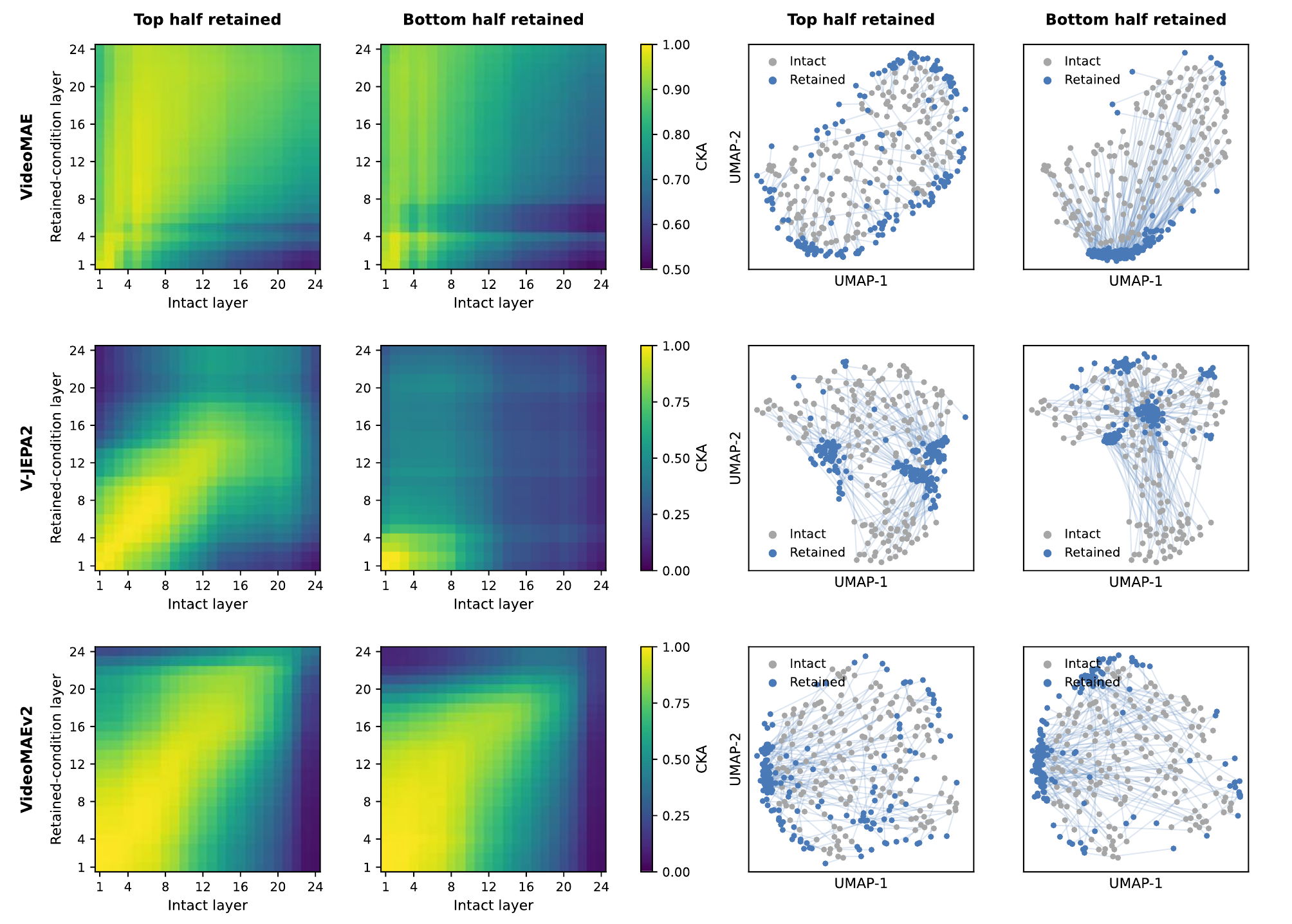}
  \caption{\textbf{Video representations with the top or bottom half of heads retained.} The first two columns show cross-layer CKA matrices comparing Intact with models retaining only the top or bottom 50\% of heads by importance. The last two columns show shared UMAP projections of final-layer video representations, with lines connecting the positions of the same video before and after intervention.}
  \label{fig:app-geometry}
\end{figure*}

Figure~\ref{fig:app-geometry} provides additional representational analyses for VideoMAE, V-JEPA 2, and VideoMAE V2.
Using the importance ranking derived from Action, we retain only the top or bottom 50\% of attention heads and compare the resulting models with the intact model.
The CKA results show that retaining high-importance heads better preserves representational similarity to Intact across multiple intermediate layers and at the final output layer.
This indicates that the differences between high- and low-importance head sets are reflected not only in the final output, but also in the evolution of representations across layers.

The shared UMAP projections further illustrate how the two interventions affect the distribution of video representations.
In VideoMAE, retaining only low-importance heads concentrates many video samples in a small region of the projection space, whereas retaining high-importance heads better preserves the original sample distribution.
In contrast, V-JEPA 2 and VideoMAE V2 exhibit sample clustering under both retention conditions, with less pronounced differences in clustering than in VideoMAE.
This suggests that clustering in two-dimensional projections alone is insufficient to explain the differences in task performance after ablating different head sets. Other representational properties and downstream computation must also be considered.

\subsection{Importance prediction}
\label{sec:cross-model-prediction}

\begin{figure*}[p]
  \centering
  \includegraphics[width=0.81\textwidth]{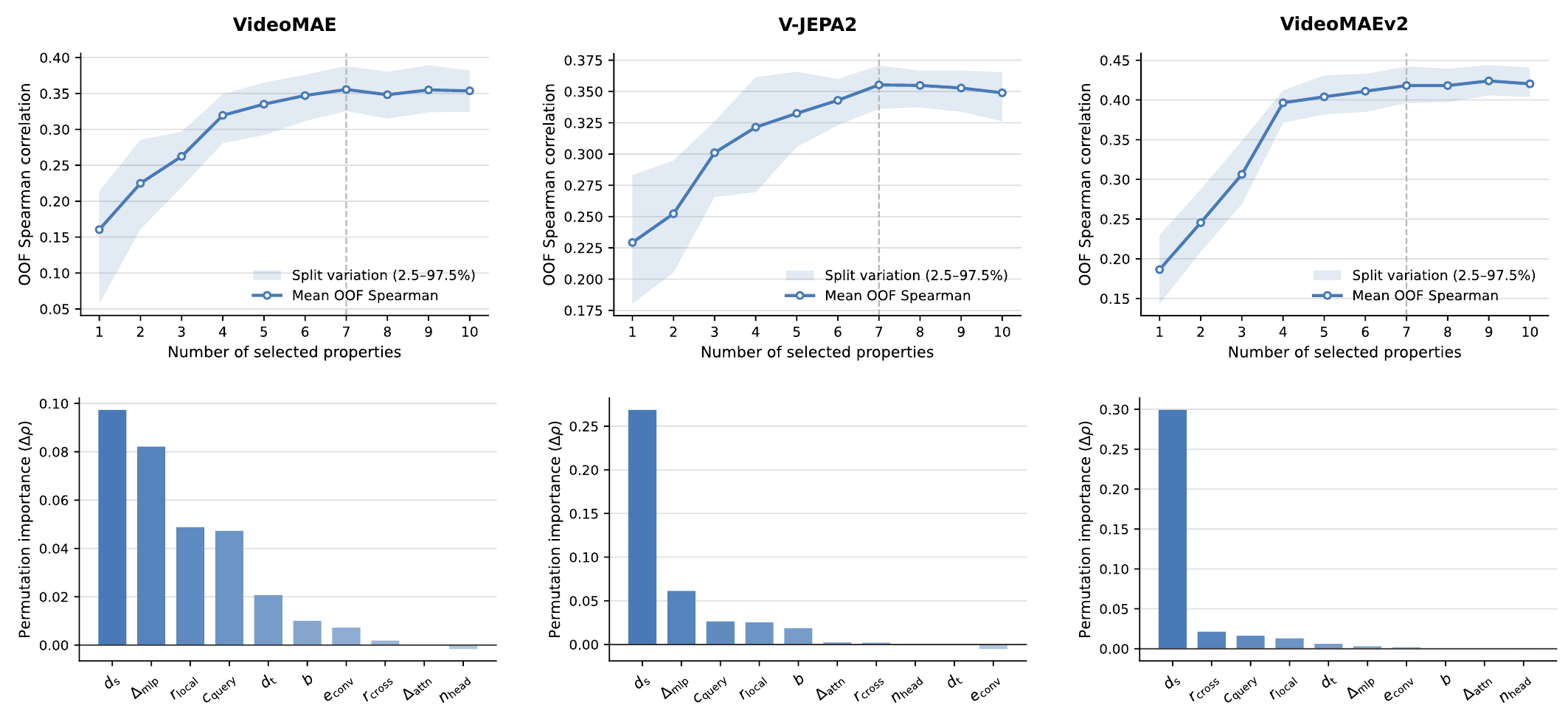}
  \caption{\textbf{Head importance prediction and property contributions across models.} The top row shows Spearman correlations between predicted and actual importance ranks for unseen heads as a function of the number of properties. Lines indicate means, and shaded regions show the 2.5th--97.5th percentile intervals across data splits. The bottom row shows permutation importance using seven properties, measured as the decrease in predictive correlation caused by randomly permuting a single property.}
  \label{fig:app-predictors}
\end{figure*}

Figure~\ref{fig:app-predictors} provides additional head importance prediction results for VideoMAE, V-JEPA 2, and VideoMAE V2.
For all three models, predictive performance generally improves as more properties are included and largely plateaus by seven properties, yielding mean Spearman rank correlations of 0.356, 0.355, and 0.418, respectively. Including additional properties produces only small fluctuations, with no consistent improvement.

Permutation importance analysis shows that the mean spatial attention distance $d_{\mathrm{s}}$ makes the largest contribution to prediction in all three models, although the contributions of the remaining properties differ. For VideoMAE, $d_{\mathrm{s}}$ and the post-MLP ablation-induced representational response $\Delta_{\mathrm{mlp}}$ contribute most, followed by the same-slice local attention ratio $r_{\mathrm{local}}$ and query pattern consistency $c_{\mathrm{query}}$.
For V-JEPA 2, $d_{\mathrm{s}}$ is the dominant predictor, followed by $\Delta_{\mathrm{mlp}}$, with smaller contributions from $c_{\mathrm{query}}$ and $r_{\mathrm{local}}$.
For VideoMAE V2, $d_{\mathrm{s}}$ contributes substantially more than any other property.
These results identify spatial attention range as a shared predictive cue across the three models, with model-dependent contributions from post-MLP representational responses and attention weight structure.

\subsection{Attention range and collective contributions}
\label{sec:cross-model-collective}

\begin{figure*}[p]
\centering
\includegraphics[width=0.92\textwidth]{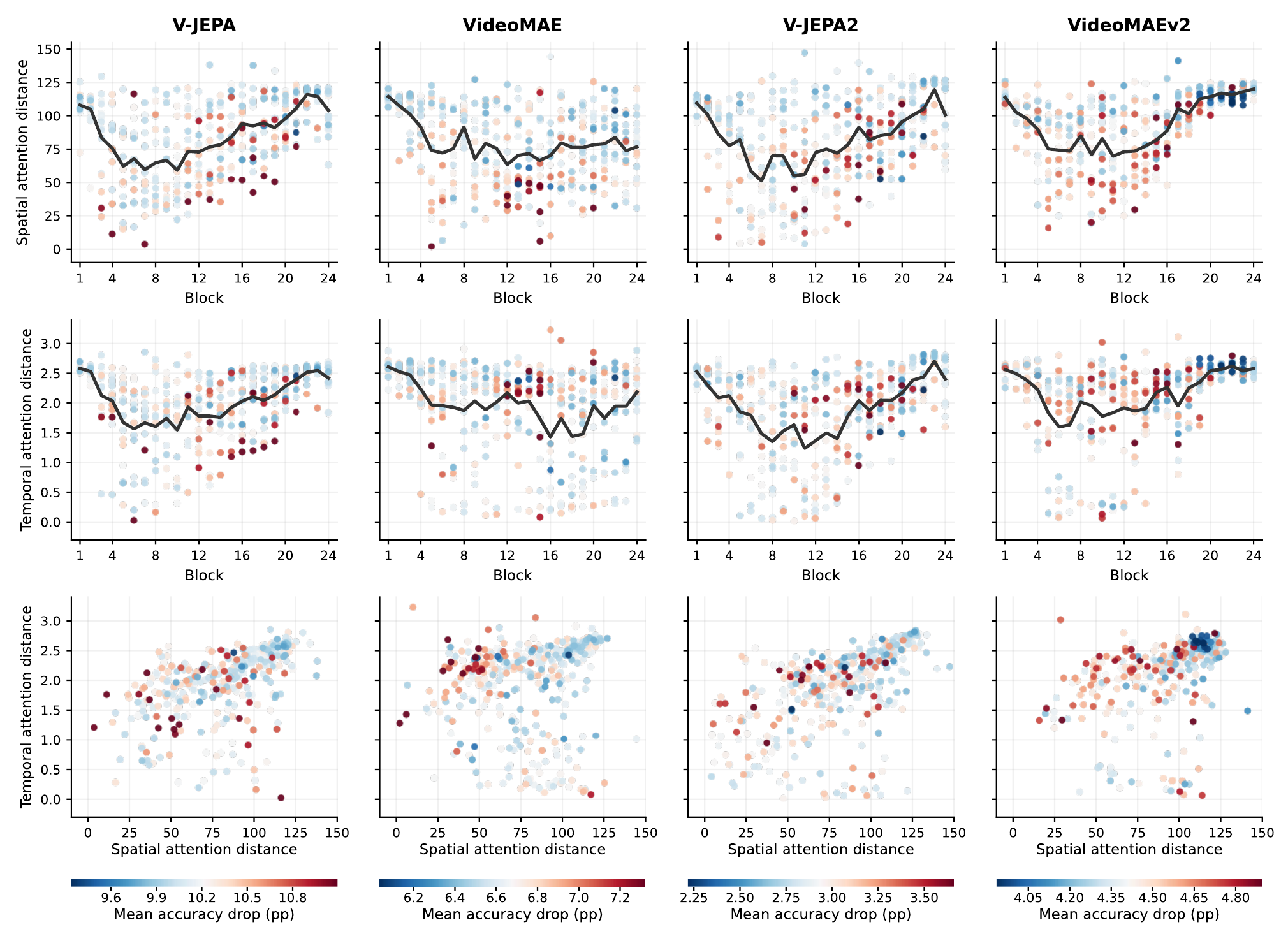}
\caption{\textbf{Spatiotemporal attention range and head importance across models.} The first two rows show the layerwise distributions of mean spatial attention distance $d_{\mathrm{s}}$ and mean temporal attention distance $d_{\mathrm{t}}$, respectively. Points represent individual heads, and lines indicate the mean within each block. The third row shows the joint distribution of the two distances. Point colors indicate head importance on Action, measured as the mean accuracy drop across random joint ablation trials containing the head.}
\label{fig:app-distance}
\end{figure*}

Figure~\ref{fig:app-distance} shows the spatiotemporal attention ranges of heads and their relationships with importance across the four models.
In each model, the mean attention distance within each block varies with network depth, with shorter distances typically occurring in the middle layers.
Attention ranges also vary substantially among heads within the same block.
Across all four models, mean spatial attention distance $d_{\mathrm{s}}$ is negatively correlated with importance on Action, indicating that heads with shorter spatial attention ranges tend to have higher importance.
In contrast, the relationship between mean temporal attention distance $d_{\mathrm{t}}$ and importance is weaker and varies across models.
These results complement the property-based importance prediction analysis in the main text and further illustrate the connection between spatial locality and head importance.

\begin{figure*}[p]
  \centering
  \includegraphics[width=\textwidth]{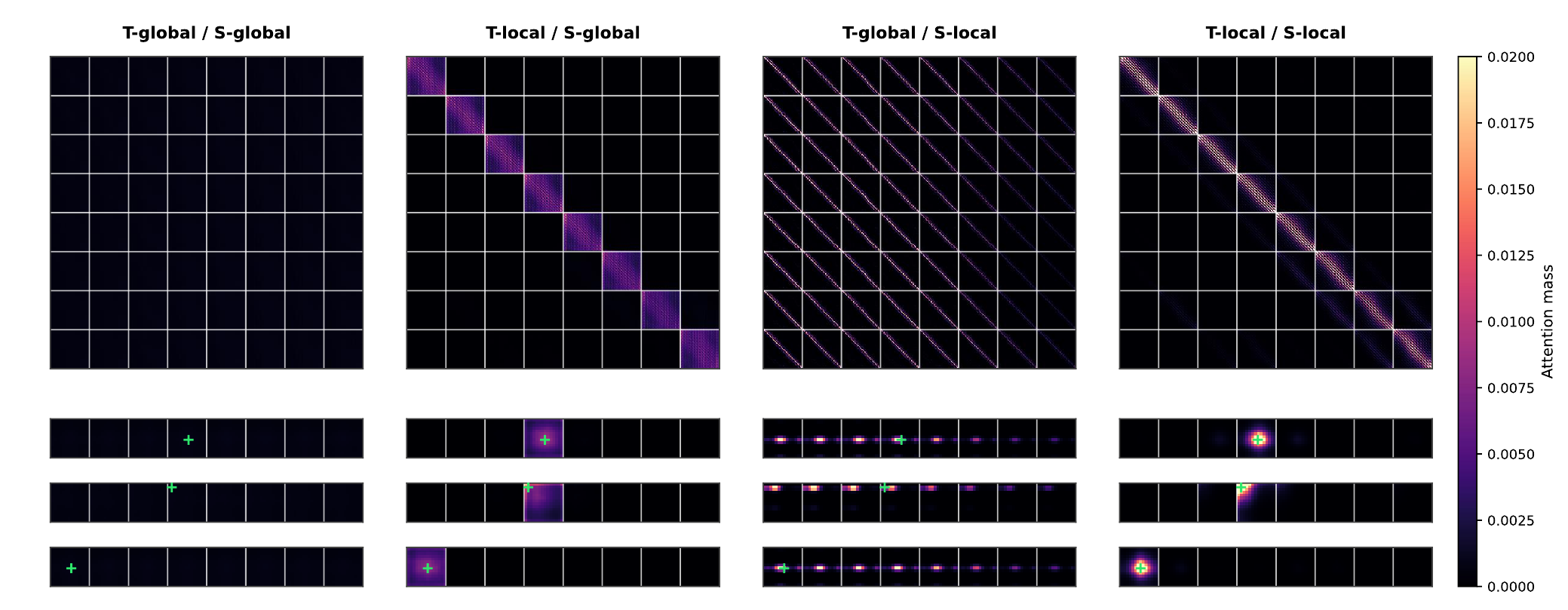}
  \caption{\textbf{Attention patterns across different spatiotemporal attention ranges.} The four columns show V-JEPA heads that are temporally global/spatially global, temporally local/spatially global, temporally global/spatially local, and temporally local/spatially local, respectively. The first row shows the full attention matrices, with queries and keys grouped by temporal slice in temporal order and spatial tokens within each slice flattened into a contiguous sequence. Grid lines mark boundaries between temporal slices. The remaining three rows show the attention distributions across temporal slices for three query positions, with green crosses marking the query positions.}
  \label{fig:app-routing-examples}
\end{figure*}

Figure~\ref{fig:app-routing-examples} uses four attention heads from V-JEPA to illustrate the attention patterns associated with different spatial and temporal attention ranges.
Temporally local heads concentrate attention mainly in the temporal diagonal blocks of the full attention matrix, while spatially local heads focus on the spatial neighborhood of the query position within the time slices they attend to.
Spatial and temporal attention ranges can combine in different ways, yielding four patterns: temporally global/spatially global, temporally local/spatially global, temporally global/spatially local, and temporally local/spatially local.
For spatially local heads, visualizations at three different query positions further show that the region of concentrated attention shifts with the query position, reflecting local aggregation around each query.

\begin{figure*}[p]
  \centering
  \includegraphics[width=0.9\textwidth]{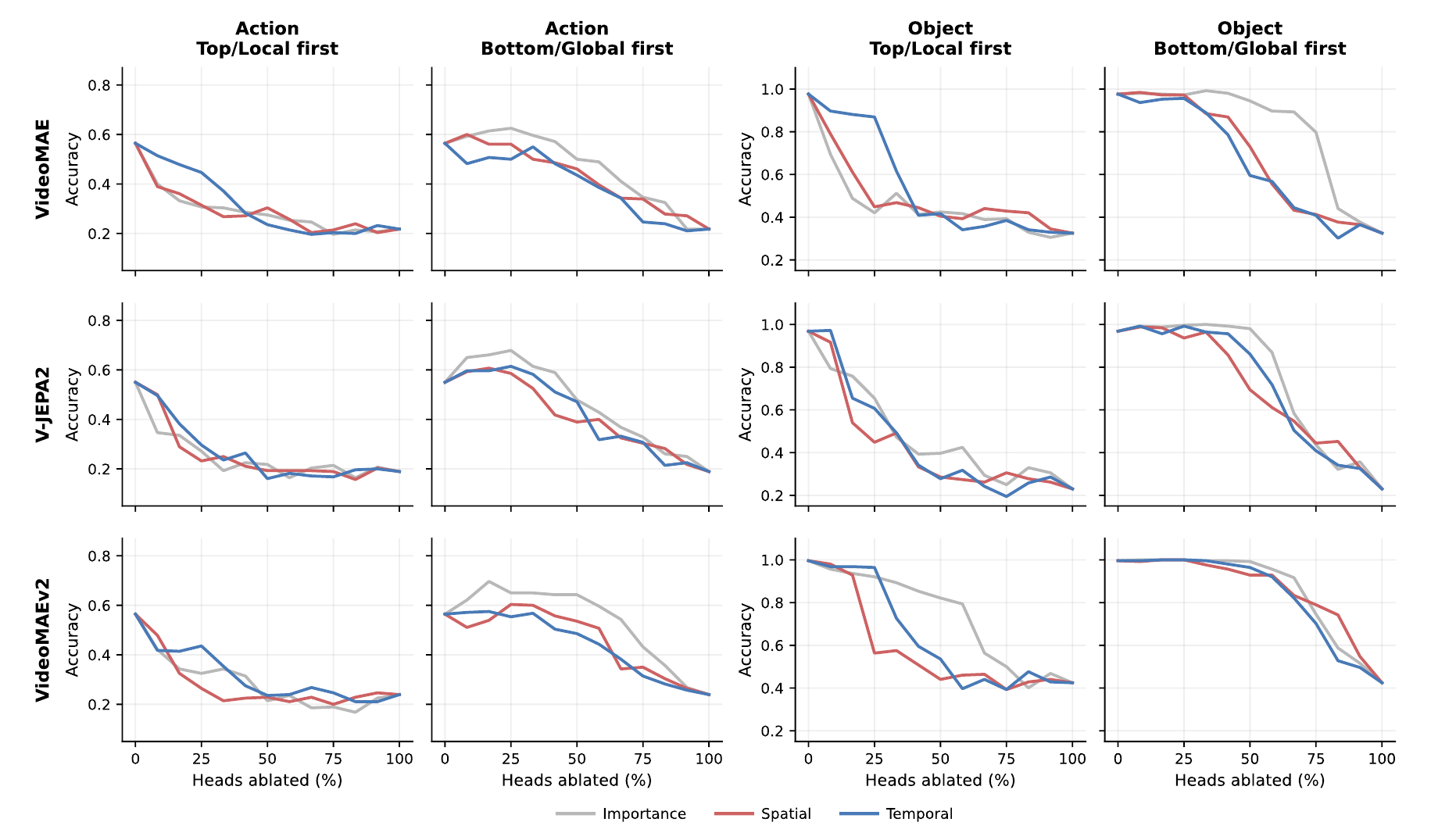}
  \caption{\textbf{Cumulative ablation of head sets across models and tasks.} Heads are ranked by importance on Action or by mean spatial or temporal attention distance. The first and third columns ablate high-importance or local heads first, while the second and fourth columns ablate low-importance or global heads first.}
  \label{fig:app-routing-ablation}
\end{figure*}

Figure~\ref{fig:app-routing-ablation} compares cumulative ablation results for the other three models, with heads ranked by importance, spatial attention distance, or temporal attention distance.
At 25\% head ablation, ablating spatially local heads first causes greater Action performance degradation than ablating high-importance heads first in V-JEPA 2 and VideoMAE V2, and comparable degradation in VideoMAE.
On Object, ablating spatially local head sets causes comparable or greater degradation in all three models.
In contrast, ablating spatially or temporally global heads first generally slows performance decline.
Together with the main-text V-JEPA experiments, these results show that jointly ablating spatially local head sets substantially degrades performance across models and tasks, and that individual head importance rankings cannot fully predict their collective functional contributions.


\clearpage

\begin{table*}[p]
  \section*{Supplementary figures and tables}
  \centering
  \small

  \caption{\textbf{Ten computational properties of attention heads.}
  Symbols, descriptions, and related work for each property.
  Detailed definitions are provided in
  Appendix~\ref{sec:head-property-definitions}.}
  \label{tab:head-properties}

  \setlength{\tabcolsep}{5pt}
  \renewcommand{\arraystretch}{1.1}
  \setlength{\heavyrulewidth}{1.2pt}
  \setlength{\lightrulewidth}{0.5pt}
  \setlength{\cmidrulewidth}{0.5pt}

  \begin{tabular}{@{}
    p{0.14\textwidth}!{\hspace{0.02\textwidth}}
    l!{\hspace{0.02\textwidth}}
    p{0.48\textwidth}!{\hspace{0.02\textwidth}}
    p{0.10\textwidth}
  @{}}
    \toprule
    \textbf{Category} &
    \textbf{Symbol} &
    \textbf{Description} &
    \textbf{References} \\
    \specialrule{0.5pt}{2pt}{1.5pt}
    \specialrule{0.5pt}{0pt}{2pt}

    \multirow[t]{4}{0.14\textwidth}{Spatiotemporal attention range}
      & $d_{\mathrm{s}}$
      & Attention-weighted mean spatial distance
      & \cite{dosovitskiy2021image} \\
      & $d_{\mathrm{t}}$
      & Attention-weighted mean temporal distance
      & \cite{bertasius2021space} \\
      & $r_{\mathrm{cross}}$
      & Fraction of attention weight assigned to other temporal slices
      & \cite{wu2022memvit} \\
      & $r_{\mathrm{local}}$
      & Fraction of attention weight within the same-slice neighborhood
      & \cite{li2022mvitv2} \\
    \midrule

    \multirow[t]{2}{0.14\textwidth}{Attention weight structure}
      & $e_{\mathrm{conv}}$
      & Deviation of the mean attention matrix from a shared kernel
      & \cite{cordonnier2020relationship} \\
      & $c_{\mathrm{query}}$
      & Consistency of relative attention patterns across query positions
      & \cite{kovaleva2019revealing} \\
    \midrule

    \multirow[t]{3}{0.14\textwidth}{Outputs and representational responses}
      & $\Delta_{\mathrm{attn}}$
      & Post-attention representational change induced by head ablation
      & \cite{chefer2021transformer} \\
      & $\Delta_{\mathrm{mlp}}$
      & Post-MLP representational change induced by head ablation
      & \cite{abnar2020quantifying} \\
      & $n_{\mathrm{head}}$
      & Norm of the head's projected output
      & \cite{kobayashi2020attention} \\
    \midrule

    Network position
      & $b$
      & Normalized depth of the block containing the head
      & \cite{raghu2021vision} \\
    \bottomrule
  \end{tabular}
\end{table*}

\clearpage

\end{document}